Auditing image datasets with GAN discriminators to protect against dirty & clean-label attacks

**John W. Belanger Smutny**
Virginia Tech University
smutnyj@vt.edu

## Abstract

Gathering enough images to train a deep computer vision model is a constant challenge. Unfortunately, collecting images from unknown sources can leave your model's behavior at risk of being manipulated by a dirty-label or clean-label attack unless the images are properly inspected. Manually inspecting each image-label pair is impractical and common poison-detection methods that involve re-training your model can be time consuming. This research uses GAN discriminators to protect a single class against mislabeled and different levels of modified images. The effect of said perturbation on a basic convolutional neural network classifier is also included for reference. The results suggest that after training on a single class, GAN discriminator's confidence scores can provide a threshold to identify mislabeled images and identify 100% of the tested poison starting at a perturbation epsilon magnitude of 0.20, after decision threshold calibration using in-class samples. Developers can use this report as a basis to train their own discriminators to protect high valued classes in their CV models.

## 1. Introduction

While navigating the Internet, safety is one thing that isn't important until it is. There are plenty of risks to such an open environment; the risk of downloading poisoned images to train deep computer vision (CV) models being one of them. Software developers and engineers designing image based machine learning (ML) models need lots of images to train and test on (1000+ for most cases). Since having thousands of images locally available for every possible object and situation is impractical, developers can use the internet to gather more. However, those images could unknowingly be mislabeled to trigger a dirty-label or altered to trigger a clean-label attack, causing your model to work in unexpected ways. Both CV classifier attacks are meant to change how the model recognizes a specific class. Dirty-label attacks involve images being mis-labeled as something different (a picture of a cat is labeled 'boat') during training, while a clean-label attack includes correctly labeled images but there are specific features ingrained into the image (sometimes hidden to the human eye, but not a CV model). At its worst, misclassification can lead to damages to people and personal property, such as making an autonomous car classify a stop sign as a garbage can, thus not stopping the vehicle and continuing into traffic. There are plenty of methods to weed out and find these poisoned images, but they often require training a model multiple times. For your most sensitive classes, why not train a full time auditor to protect against such a targeted attack? Why not use a Generative Adversarial Network (GAN) discriminator?

A traditional GAN requires training two models; a generator to make images and a discriminator to judge images. Normally, the discriminator judges the generator's creations to answer: "is this image real or fake". However, this work changes the discriminator's question to: "Is this image what it ought to be?". By training a discriminator to pick out the normal features of a specific class, developers could use them as initial auditors for anomaly detection for that class. In practice, whenever developers integrate a new set of images from unknown sources, the discriminator could be used to protect against poisoned images (see the flow diagram below for an example implementation). It is realistic to believe that GAN discriminators can act as these first auditors, because during training the discriminator only learns from unpoisoned images, thus any mislabeled or poisoned images with added features should place that sample's features outside of the discriminator's understanding of the class.

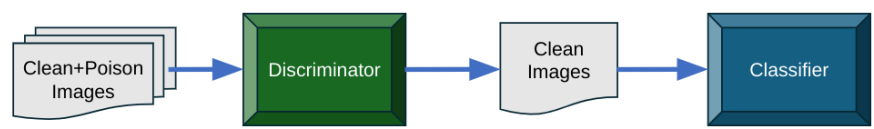

*Figure 1.1: Flow diagram of the use case for using GAN discriminators to find poisoned or anomaly images. A new dataset is fed into the discriminator, the model identifies certain images for review, the remaining images (ideally without poison) are used for classifier training & testing.*

## 2. Related Research

Federico Di Mattia et al. outlined various ways GANs could be used for anomaly detection [Di Mattia]. They describe various architectures and methods training GAN discriminators to best detect when there is an outlier sample (AKA: a mislabeled/poisoned sample). This project experiments with Di Mattia's AnoGAN architecture to train a GAN discriminator for poisoned images using only positive samples of the desired class, while expanding their idea to cybersecurity defense for the desired class. Along with providing an outline of the effect of any missed poison on a simple classifier model, the idea of training discriminators on multiple classes to improve poison detection scope was investigated but not included in this report.

Most other papers explore how GANs can be used to generate poison, not protect against poisoned images. Academic papers can be found that describe components of GANs being poisoned [Jin] or using GANs to generate poisoned samples [Zhao]. Existing poison-detection research has identified several methods of identifying poisoned image samples (clean or augmented). Methods such as neuron pruning, Bilevel optimization, and meta-sift have been documented to remove poisoned neurons, train against poisoning, and detect poisoned samples respectively. The meta-sift method of poison detection in particular requires the user to retrain their models several times. Another method of poison detection compares the entropy of individual samples, higher loss indicating a poisoned sample.

## 3. Methodology

To explore the feasibility of using a GAN discriminator to audit new datasets, the MNIST handwritten digits dataset is used for dirty & clean-label attacks for each digit. In both types of attacks the discriminator judges if the given image is realistic for its trained class. This paper's hypothesis is that mislabeled images and any added perturbations pushes the image out of the learned feature space. The experiments are repeated on a discriminator trained for each MNIST digit. No results regarding GAN discriminators trained on multiple classes is provided in this paper. In practice, having a discriminator trained on multiple classes would increase the scope of protection, but most likely at the cost of protection effectiveness. Effects of any missed poison samples with this studied perturbation on a basic classifier is shown in appendix A3 for context. For this research, a GAN discriminator trained to judge one specific MNIST digit is referred to as *discriminator-x*. Where *x* is the digit in which the discriminator is trained to judge as real or fake. All other MNIST digits not *x* are considered out-of-class. Also, 'in-class' refers to the 'x' class that discriminator-x is trained to identify.

First, discriminators are tested with samples of all possible MNIST classes but are re-labeled for the discriminator's trained digit. This experiment involves recording the discriminator's confidence scores for 500 images matching their trained class and 4500 images outside of their trained class. The goal is to find differences in confidence score distributions for each digit class. It is expected that the confidence distribution (max, min, mean) for the in-class images are higher than each of the out-of-class images. Second, discriminators are tested using only digits of the same class as the trained on digit, but with some test samples being poisoned by perturbations at different epsilon values. Before testing, the discriminator's decision threshold deciding if an image is poison or clean is calibrated to the average confidence score of a separate set of epsilon 0.0 in-class images. During testing, 2000 in-class images are used, where 1000 of those images contain some level of perturbation. As the magnitude of perturbation increases, it is expected that the rate of true-positives (predicted poison & actual poison) will also increase.

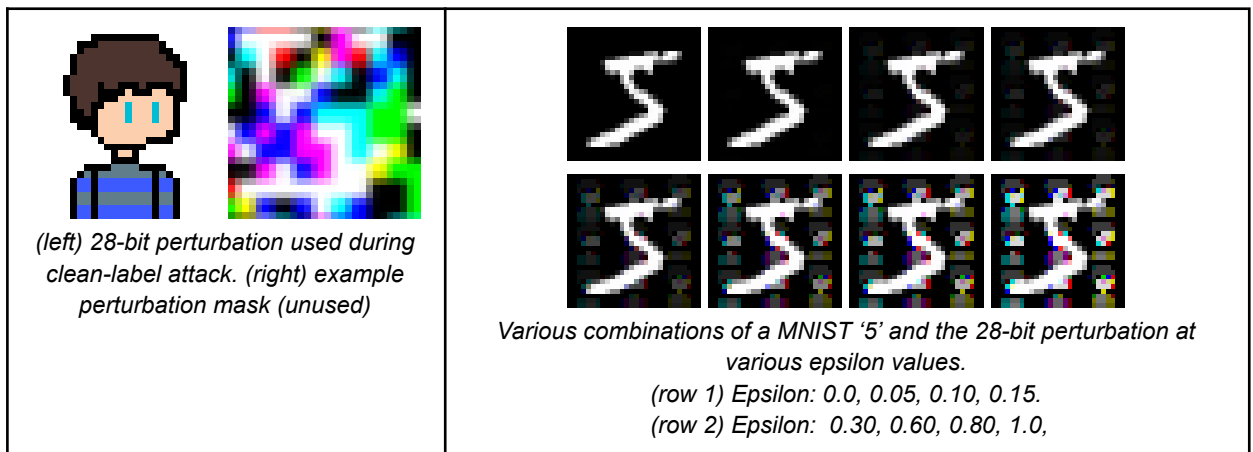

*(left) 28-bit perturbation used during clean-label attack. (right) example perturbation mask (unused)*

*Various combinations of a MNIST '5' and the 28-bit perturbation at various epsilon values.*
*(row 1) Epsilon: 0.0, 0.05, 0.10, 0.15.*
*(row 2) Epsilon: 0.30, 0.60, 0.80, 1.0,*

*FIgure 3.1: (left) examples of perturbations that can be hidden into images during a clean-label attack. (right) examples of a perturbation at different magnitudes of epsilon.*

### 3.2. Model architectures

Across all of the experiments, three ML models are trained (GAN generator, GAN discriminator, and a CNN classifier) but only the GAN discriminator and CNN classifier are used. The GAN system composed of a generator and discriminator have different architectures for their respective roles. The generator is a 4x layer model consisting of a Dense layer and three Conv2DTranspose layers to upscale a single neuron to a 28x28x1 mock MNIST image for judgement. The discriminator is a CNN model consisting of 4x Convolution-MaxPooling-LeakyReLU-Dropout layers with a linear output activation function. No batch normalization layers are used since they seemed to hinder GAN training and a linear activation function (rather than sigmoidal common to binary classifiers) is used to better document absolute discriminator confidence (better set a decision threshold). It wasn't uncommon during testing that the discriminator would have a non-linear output of 0.99 confidence for most images (correct or mislabeled). Such a common confidence output makes it difficult to tune the discriminator's decision threshold. The mock CNN classifier model is a traditional CNN with three Convolution-MaxPool-BatchNormalize layers.

|  | *Classifier* | *GAN: Generator* | *GAN: Discriminator* |
|---|---|---|---|
| *Learning rate (decay)* | 1e-4 (95% per epoch) | 1e-5 (97% per epoch) | 1e-5 (97% per epoch) |
| *Loss function* | Categorical Cross Entropy | Bi-cross Entropy | Bi-cross Entropy |
| *Optimizer* | Adam (0.9, 0.999) | - | - |
| *Batch size* | 16 | 32 | 32 |
| *Epochs* | 12 | 75 | 75 |
| *Kernel regularization* | L2 (1e-5) | L2 (1e-5) | L2 (1e-5) |
| *Number of training samples* | 500 per class | 1000 of one class | 1000 of one class |
| *Number of testing images* | 250 per class | - | - |

*Figure 3.2.1: High level training settings for each model training.*

## 4. Results

The key to GAN discriminators being used as an auditor against dirty & clean-label attacks is trust. Trust in the discriminator's ability to find all of the bad images and not label bad images as clean images. As documented in appendix A3 for the clean-label attack case, having a basic CNN train on only a few training poison images can significantly impact the classifier's performance. As such, this paper focuses on determining how well these single class trained discriminators can accurately detect images mislabeled or poisoned with perturbations. All results in the subsequent sections are reflective of GAN discriminators trained for a specific target class unless otherwise stated.

### 4.1. Detecting mis-labeled images (dirty-label attack)

Based on the confidence score distributions between in-class and out-of-class images, it is possible for a discriminator to locate and reject mislabeled classes. Between the 10 MNIST classes, most GAN discriminator performance either had 1) a clear separation, 2) separation between all but a few classes, or rarely that 3) the in-class distribution was not separate from out-of-class distributions. As seen in figure 4.1.1 for example, the discriminator-0 model (left) clearly has a higher confidence for in-class '0' images compared to the confidence distributions of other digits. Therefore, the '0' class can easily be protected against a dirty-label attack, since a developer can place a confidence threshold that will flag any image with a lower than expected confidence score as a potentially mislabeled image for human review. Discriminator-5 (middle) does have distribution separation but not as clear as discriminator-0. This behavior is the most common for the MNIST dataset. As the last example, discriminator-4 (right) shows that a discriminator cannot be blindly used for mislabel protection. Out-of-class images for '1' or '9' have higher confidence distributions than the in-class images and thus any image of a '1' or '9' would pass the discriminator filter. Performance for discriminators for all digits are shown in appendix A1.

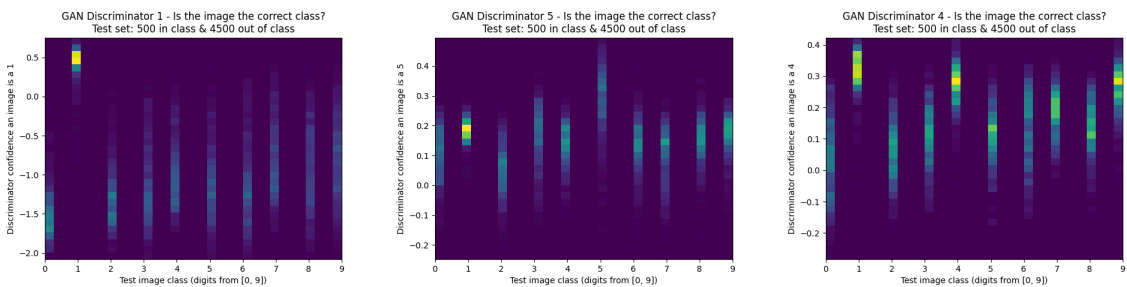

*FIgure 4.1.1: Three example GAN discriminator confidence score distributions for images of the class the discriminator was trained with versus not trained with. Discriminators for class 1 & 5 show high and mild score distributions allowing for possible filtering. DIscriminator for class 4 cannot separate images of class 4 vs outside classes 1 or 9.*

## 4.2. Detecting poison images (clean-label attack)

Based on confusion matrix performance of discriminators labeling if an image has poison, it is confirmed that discriminators can be used for poison detection after the discriminator's decision threshold has been calibrated. This experiment utilizes confusion matrices to quantify discriminator performance (positive=*poison*, negative=*clean*). Ideal performance is that the discriminator finds all poison samples (true-positives) while not predicting any poison as *clean* (false-negatives) and minimizing clean samples being predicted as *poison* (false-positive). In these mislabel situations, there is a low cost to labeling clean images as poison since the only consequence is having more images to review or remove by a human. This is an inconvenience compared to the high cost of labeling a poisoned image as clean since, as detailed in appendix A3, each trained on poison image can cause significant changes to model performance.

|  | *Predicted* | *Actual* | *Cost of mislabeling* |
|---|---|---|---|
| *True-Positive (TP)* | poison | poison | - |
| *True-Negative (TN)* | clean | clean | - |
| *False-Positive (FP)* | poison | clean | low |
| *False-Negative (FN)* | clean | poison | high |

*Figure 4.2.1: Truth table detailing the labeling method of GAN discriminators judging each newly found training image. Overly-cautious models eliminate FN samples to protect classifiers at the cost of losing clean training data.*

As seen in the selected example stack bar charts in figure 4.2.2., all decision threshold calibrated discriminators followed similar performance of showing that as poison epsilon increases the rate of true-positive and false-positive identification increases, while the false-negative rate increases, and true-negative rate remains constant. Some digit discriminators showed a quicker convergence to 0% false-negatives (predicted clean but poison), but all reached 0% at about 0.2 or 0.3 epsilon. One consequence of having a calibrated decision threshold is catching stealthier poison at the cost of having a consistent number of false-positives (predicted poison but clean). However, calibration is substantial since non-calibrated discriminators struggle to identify any poison until it reaches a magnitude of 0.1. All of the stack plots for discriminator-x models can be seen in appendix A2.

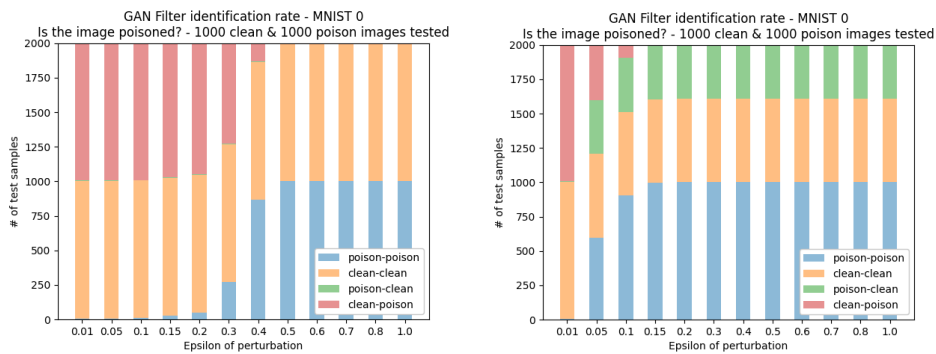

*Figure 4.2.2: stack plots for a discriminator-0 performance recognizing images of class 0 as having poison or not. (left) Confusion bar chart with the decision threshold set to 0.0. (right) Confusion bar chart with the decision threshold calibrated to the average confidence values of images of class 0 without perturbations (epsilon 0.0).*

## 5. Discussion

This research's original hypothesis believed that GAN discriminator models can learn a class well enough to defend against dirty-label and clean-label attacks. Results in this paper support that discriminators can plausibly defend against mislabeled images (depending on the class and adjacent classes) and the results confirmed that discriminators can find image perturbations (consult your project requirements for maximum allowed perturbations). For most MNIST classes, trained discriminators had the highest confidence on their in-class test images, but other classes that have similar features (like 1/4/9) did not have enough separation between confidence distributions to have the discriminator be the sole protection against dirty-label attacks. On the other hand, for all MNIST classes, calibrated discriminators could find 100% of in-class images that had epsilon magnitude 0.3 poison perturbations. Based on the use case's requirements adjusting the decision threshold higher can find even lower magnitude poison or not misclassify valid clean images if set lower. Based on the results for these two hypotheses, this author recommends that GAN discriminators can be used as a first layer of poison detection to supplement other poison detection methods. It is not recommended to have discriminators as the only defense structure.

### 5.1. Other uses of GAN discriminators (Data quality)

Even if GAN discriminators are not sensitive enough to find perturbation poison lower than epsilon 0.10 during a clean-label attack, it is possible that a trained discriminator can be used as a method to determine data quality. Given enough training samples, a discriminator's judgement on known clean data can help determine training samples that fall within a known feature space. Going back to the original discriminator question of "is this image a real picture of class x?". Given a new dataset, some samples might be low quality, blurry, out of focus, or have the desired class obscured. These samples are valid, but may harm a classifier's training by creating unhelpful learning gradients. Compared to ideal images of that class, a discriminator-x wouldn't be nearly as confident in labeling them as a part of that class. Developers could use these insights to see if images labeled as 'fake' are of good enough quality to train off of, even if it doesn't have any poison. This reduces the burden on developers to find bad samples and can improve end model performance by not training on wrongly labeled or images with harmful non-poisonous features. However, over-use of this method can cause a resulting dataset to not have enough variety of class variations to avoid model overfitting.

### 5.2. Way to optimally set your decision threshold

As previously stated in section 4.2., setting the perturbation detection GAN discriminator's decision threshold to a value other than 0.0 can yield substantial benefits. In this report, the decision threshold is set to the average confidence value of a separate set of in-class images that did not have any perturbations (epsilon 0.0). Much like figure 4.1.1.'s distribution curves, the in-class based threshold establishes 'normal' performance so that any perturbations that cause a change in performance can be better detected. For future extensions of this research into poison detection, the decision threshold could be more accurately set by using the ROC curve for the smallest allowed perturbation. ROC curves can be modified to specifically show the relationship between TP and FN (the two highest cost labels) to see at what threshold yields zero FN samples in the test set for the desired perturbation level.

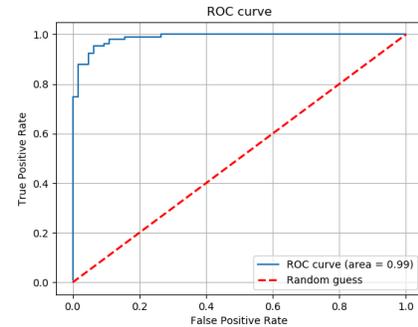

### 6. Conclusion

ML models have a constant need for high quality data, but that need doesn't mean developers need to compromise their model's security for more samples. These experiments provide evidence that training GAN discriminators for high priority classes can provide an additional tool to stop dirty-label & clean-label attacks before they happen. Single class trained discriminators can learn that class's usual set of features and can spot mislabeled and perturbed images trying to disrupt the model's behavior. Future researchers can apply these models and guidelines to their own use cases and improve upon this work by using improved GAN architectures and loss functions (such as those proposed by Federico Di Mattia and the Wasserstein loss function). Happy hunting.

**References**

Mettia, Federico Di, Galeone, Paolo., Simoni, Michele De., Ghelfi, Emanuele. A Survey on GANs for Anomaly Detection. ArxIv, 2019. Vol. 1906.11632.

Jin, R., Li, X. (2022). Backdoor Attack is a Devil in Federated GAN-Based Medical Image Synthesis. In: Zhao, C., Svoboda, D., Wolterink, J.M., Escobar, M. (eds) Simulation and Synthesis in Medical Imaging. SASHIMI 2022. Lecture Notes in Computer Science, vol 13570. Springer, Cham. https://doi.org/10.1007/978-3-031-16980-9_15

Pixelart. Pixel art drawing tool. Bryan Ware. Utah. "A safe social platform for everyone. Create beautiful pixel art, share, collaborate, shop and more!". Accessed 12/04/2024. https://www.pixilart.com/

Raichur, Nisha. "Fast-Gradient-Signed-Method-FGSM". Commit Hash e6d687c, May 19th, 2020. Tool to add Fast Gradient poison to images. https://github.com/nisharaichur/Fast-Gradient-Signed-Method-FGSM

Zhao, B., & Lao, Y. (2022). CLPA: Clean-Label Poisoning Availability Attacks Using Generative Adversarial Nets. Proceedings of the AAAI Conference on Artificial Intelligence, 36(8), 9162-9170. https://doi.org/10.1609/aaai.v36i8.20902

# Appendix

## A1) Dirty-Label attack distribution plots

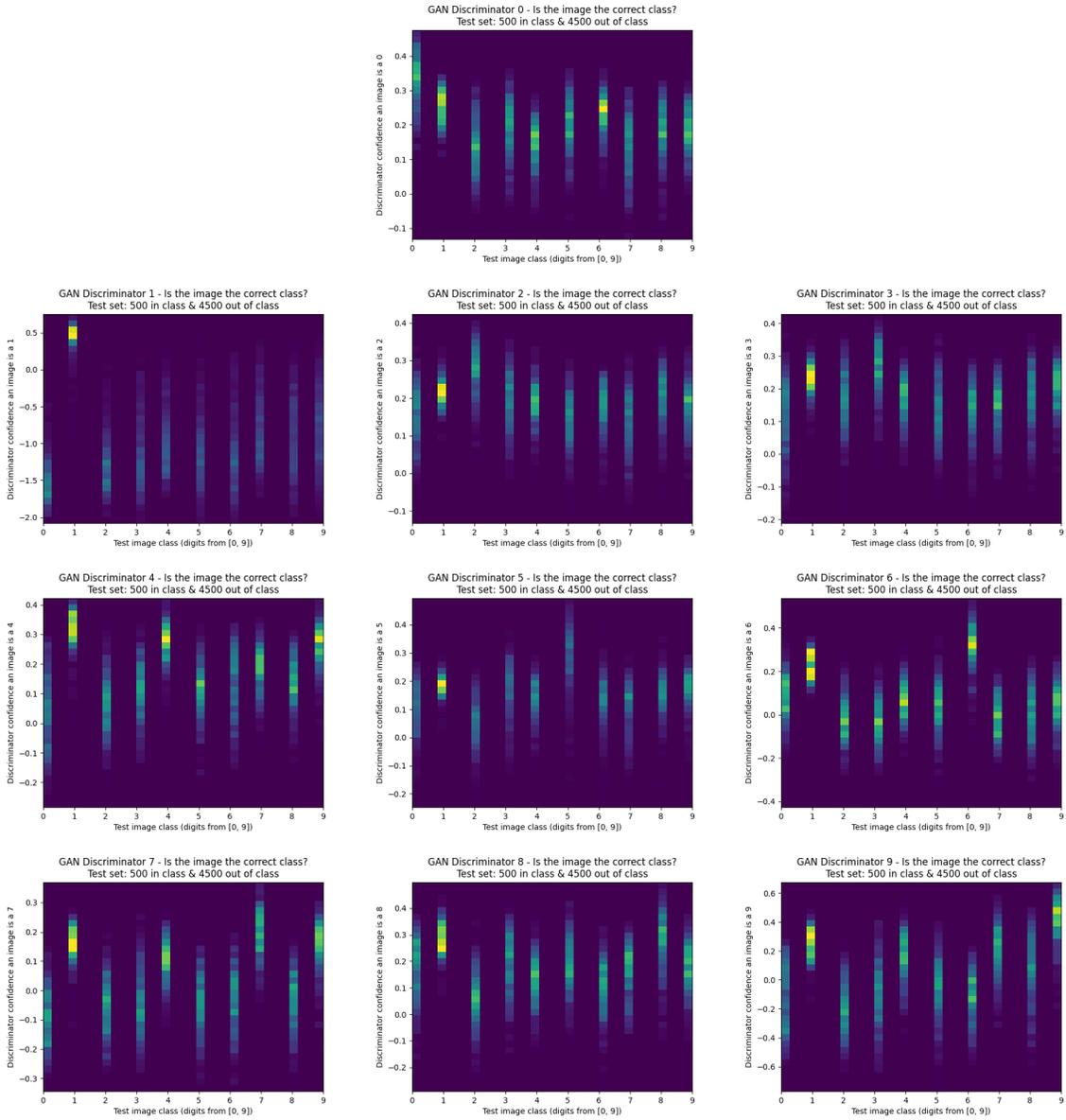

## A2) Clean-label attack confusion matrix stack plots

Stack plots representing confusion matrix binary decisions on if a test image contains poison.
- Positive = image has poison
- Negative = image does not have poison

## A2a) Stack plots when the decision threshold is calibrated

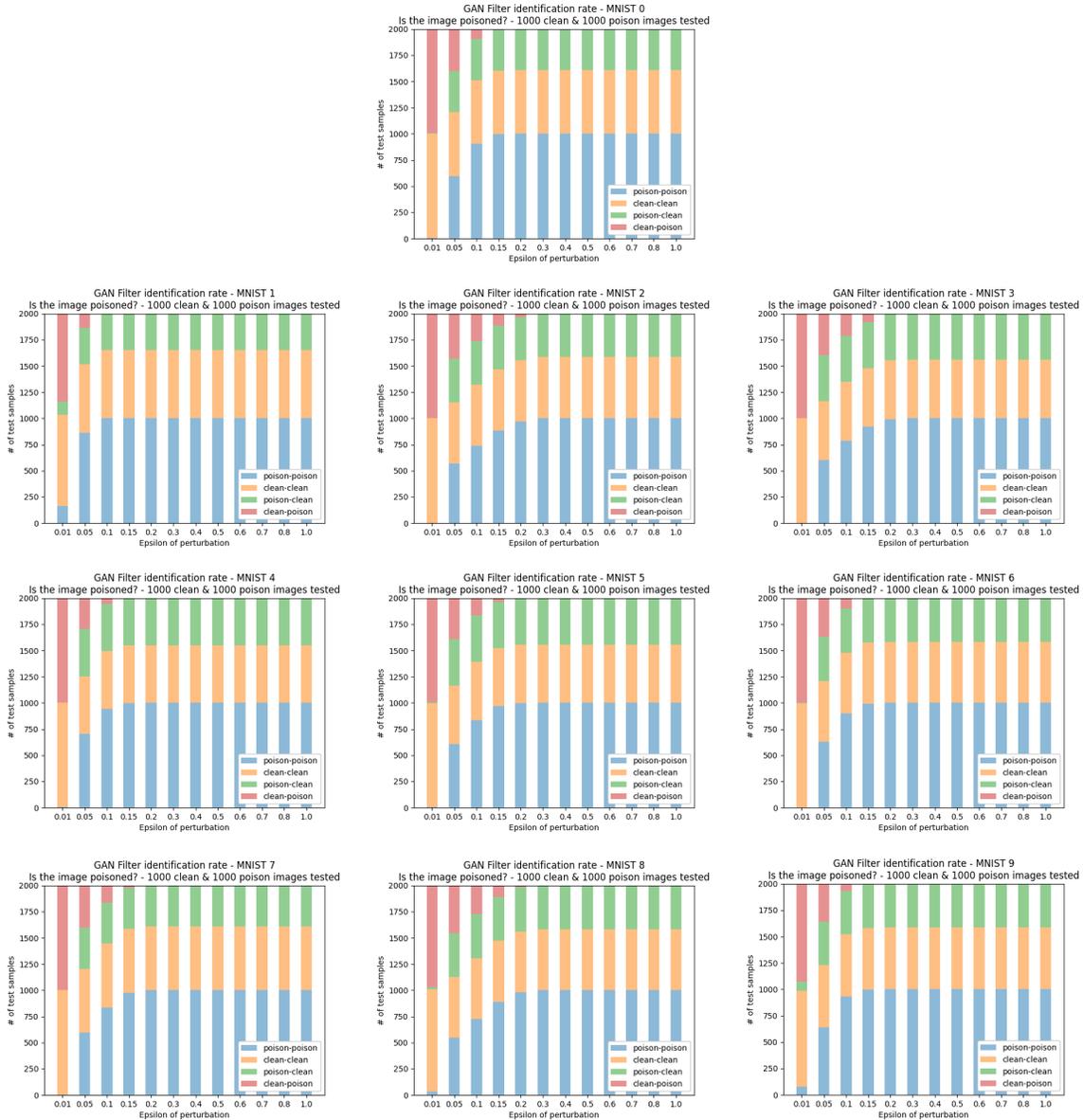

**A2b) Stack plots when the decision threshold is set to 0.0**

## A3) Effects of poisoned samples with this paper's perturbation on a CNN classifier

To better put in context the consequences of any missed poisoned training samples, a simple deep CNN model can be trained to understand how powerful poison can be. It is well documented that training on few poisoned samples can have significant impacts on test accuracy, but can that behavior be repeated? As a part of this research, a four-layer CNN multi-class classifier (CNN-max_sample-batch_renorm) is trained several times using different amounts of poisoned images and at different levels of perturbation. The average 1) test accuracy overall, 2) non-poisoned class test accuracy, 3) poisoned class test accuracy, and 4) attack success rate (ASR) are recorded and plotted in figures below. Final results are averages of three runs for each {epsilon, number of trained on poison} pairs. Only the max-min and discriminator significant perturbations are plotted for cleanliness. These results are only valid for the perturbation used in this paper. Effects of any other perturbation would need to be further tested.

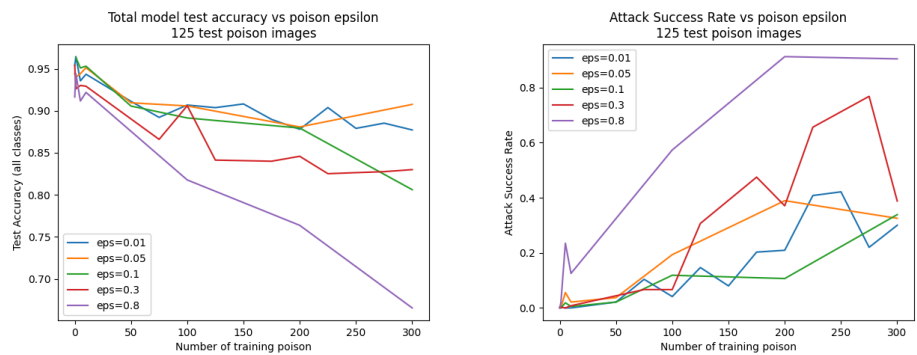

Figure A3.1: (left) Overall test accuracy of the model and (right) attack success rate with increasing number of trained on poison for various perturbations. Shows that with more poison and higher perturbations, model accuracy decreases and attack success rate increases (even with 0.01 magnitude perturbation).

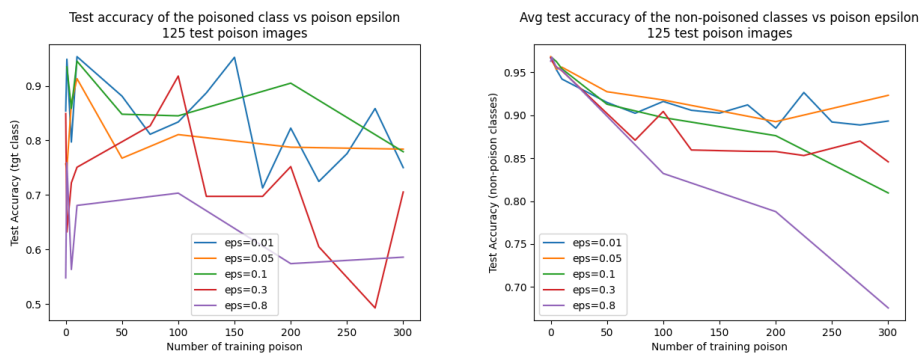

Figure A3.2: (left) Test accuracy of target class and (right) test accuracy of other classes with increasing number of trained on poison for various perturbations. Shows that with more poison and higher perturbations, poison accuracy variably decreases, but the accuracy of other classes steadily decreases.